\documentclass[10pt,twocolumn,letterpaper]{article}

\usepackage{cvpr}
\usepackage{times}
\usepackage{epsfig}
\usepackage{graphicx}
\usepackage{amsmath}
\usepackage{amssymb}
\usepackage{algorithm}
\usepackage{algpseudocode}

\usepackage{etoolbox}
\makeatletter
\patchcmd{\maketitle}
 {\def\@makefnmark}
 {\def\@makefnmark{}\def\useless@macro}
 {}{}
\makeatother

\usepackage{multirow}
\usepackage{graphicx}
\usepackage{makecell}
\usepackage{url}
\def\Figref#1{Fig.~\ref{#1}}
\def\Secref#1{Sec.~\ref{#1}}
\def\Algref#1{Alg.~\ref{#1}}
\def\Tabref#1{Tab.~\ref{#1}}
\def\Eqref#1{Eq.~\eqref{#1}}
\usepackage{amsmath,amsfonts,bm}

\def\vh{{\bm{h}}}

\def\vo{{\bm{o}}}

\def\mA{{\bm{A}}}

\def\mI{{\bm{I}}}

\def\mW{{\bm{W}}}

\def\gA{{\mathcal{A}}}

\def\gG{{\mathcal{G}}}

\def\gT{{\mathcal{T}}}

\def\gW{{\mathcal{W}}}

\def\sD{{\mathbb{D}}}
\def\sE{{\mathbb{E}}}
\def\sF{{\mathbb{F}}}
\def\sR{{\mathbb{R}}}

\usepackage[breaklinks=true,bookmarks=false]{hyperref}

\cvprfinalcopy 

\def\cvprPaperID{80} 

\ifcvprfinal\fi
\begin{document}

\title{Searching for A Robust Neural Architecture in Four GPU Hours}

\author{Xuanyi Dong$^{1,2}$,\thanks{This paper was accepted to the IEEE CVPR 2019.}\thanks{Part of this work was done when Xuanyi Dong was a research intern with Baidu Research.}Yi Yang$^{1}$\thanks{Corresponding author: Yi Yang.}\\
$^{1}$University of Technology Sydney\hspace{4mm}$^{2}$Baidu Research \\
{\tt\small xuanyi.dong@student.uts.edu.au},~{\tt\small yi.yang@uts.edu.au}
}

\maketitle
\thispagestyle{empty}

\begin{abstract}

Conventional neural architecture search (NAS) approaches are based on reinforcement learning or evolutionary strategy, which take more than 3000 GPU hours to find a good model on CIFAR-10.
We propose an efficient NAS approach learning to search by gradient descent.
Our approach represents the search space as a directed acyclic graph (DAG). This DAG contains billions of sub-graphs, each of which indicates a kind of neural architecture.
To avoid traversing all the possibilities of the sub-graphs, we develop a differentiable sampler over the DAG.
This sampler is learnable and optimized by the validation loss after training the sampled architecture.
In this way, our approach can be trained in an end-to-end fashion by gradient descent, named Gradient-based search using Differentiable Architecture Sampler (GDAS).
In experiments, we can finish one searching procedure in four GPU hours on CIFAR-10, and the discovered model obtains a test error of 2.82\% with only 2.5M parameters, which is on par with the state-of-the-art.
Code is publicly available on GitHub: \url{https://github.com/D-X-Y/NAS-Projects}.
\end{abstract}

\section{Introduction}

Designing an efficient and effective neural architecture requires substantial human effort and takes a long time~\cite{cho2014learning,dong2017more,dong2018sbr,he2016deep,huang2017densely,krizhevsky2012imagenet,szegedy2015going,zhu2017bidirectional}.
Since the birth of AlexNet~\cite{krizhevsky2012imagenet} in 2012, human experts have conducted a huge number of experiments, and consequently devised several useful structures, such as attention~\cite{cho2014learning} and residual connection~\cite{he2016deep}.
However, the infinite possible choices of network architecture make the manual search unfeasible~\cite{baker2017designing}.
Recently, neural architecture search (NAS) has increasingly attracted the interest of researchers~\cite{baker2017designing,chen2018searching,kandasamy2018neural,Liu_2018_ECCV,luo2018neural,pmlr-v80-pham18a,zoph2017NAS}.
These approaches learn to automatically discover good architectures. They can thus reduce the labour of human experts and find better neural architectures than the human-invented architectures.
Therefore, NAS is an important research topic in machine learning.

\begin{figure}[t!]
\begin{center}
\includegraphics[width=\linewidth]{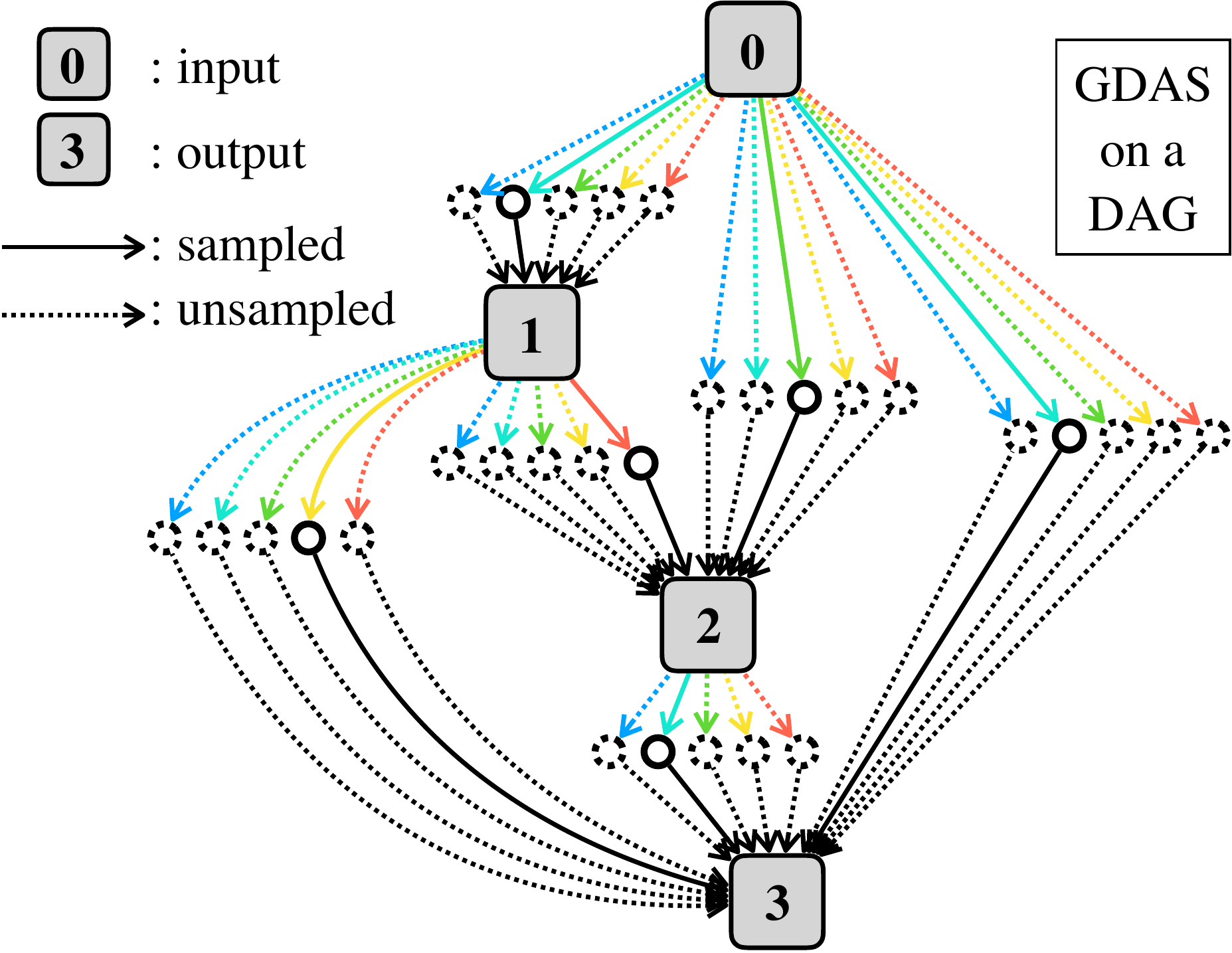}
\end{center}
\caption[Captioning]{
We utilize a DAG to represent the search space of a neural cell.
Different operations (colored arrows) transform one node (square) to its intermediate features (little circles). Meanwhile, each node is the sum of the intermediate features transformed from the previous nodes.
%
As indicated by the \textbf{solid} connections, the neural cell in the proposed GDAS is a sampled sub-graph of this DAG.
Specifically, among the intermediate features between every two nodes, GDAS samples one feature in a differentiable way.
}
\label{fig:framework}
\end{figure}

Most NAS approaches apply evolutionary algorithms (EA)~\cite{real2019regularized,liu2018hierarchical,real2017large} or reinforcement learning (RL)~\cite{zoph2017NAS,Zoph_2018_CVPR,bello2017neural} to design neural architectures automatically.
In both RL-based and EA-based approaches, their searching procedures require the validation accuracy of numerous architecture candidates, which is computationally expensive~\cite{Zoph_2018_CVPR,real2019regularized}.
For example, the typical RL-based method utilizes the validation accuracy as a reward to optimize the architecture generator~\cite{zoph2017NAS}.
An EA-based method leverages the validation accuracy to decide whether a model will be removed from the population or not~\cite{real2017large}.
These approaches use a large amount of computational resources, which is inefficient and unaffordable.
This motivates researchers to reduce the computational cost.
%

In this paper, we propose a \textbf{G}radient-based searching approach using \textbf{D}ifferentiable \textbf{A}rchitecture \textbf{S}ampling (GDAS).
It can search for a robust neural architecture in four hours with a single V100 GPU.
GDAS significantly improves efficiency compared to the previous methods.
We start by searching for a robust neural ``cell'' instead of a neural network~\cite{zoph2017NAS,Zoph_2018_CVPR}.
A neural cell contains multiple functions to transform features, and a neural network consists of many copies of the discovered neural cell~\cite{Liu_2018_ECCV,Zoph_2018_CVPR}.
\Figref{fig:framework} illustrates our searching procedure in detail.
We represent the search space of a cell by a DAG.
Every grey square node indicates a feature tensor, numbered by the computation order.
Different colored arrows indicate different kinds of operations, which transform one node into its intermediate features.
Meanwhile, each node is the sum of the intermediate features transformed from the previous nodes.
During training, the proposed GDAS samples a sub-graph from the whole DAG, indicated by solid connections in \Figref{fig:framework}.
In this sub-graph, each node only receives one intermediate feature from every previous node.
Specifically, among the intermediate features between every two nodes, GDAS samples one feature in a differentiable way.
In this way, GDAS can be trained by gradient descent to discover a robust neural cell in an end-to-end fashion.

The fast searching ability of GDAS is mainly due to the sampling behavior.
A DAG contains hundreds of parametric operations with millions of parameters.
Directly optimizing this DAG~\cite{liu2019darts} instead of sampling a sub-graph leads to two disadvantages.
First, it costs a lot of time to update numerous parameters in one training iteration, increasing the overall training time to more than one day~\cite{liu2019darts}.
Second, optimizing different operations together could make them compete with each other. For example, different operations could generate opposite values.
The sum of these opposite values tends to vanish, breaking the information flow between the two connected nodes and destabilizing the optimization procedure.
To solve these two problems, the proposed GDAS samples a sub-graph at one training iteration.
As a result, we only need to optimize a part of the DAG at one iteration, which accelerates the training procedure.
Moreover, the inappropriate competition is avoided, which makes the optimization effective.

In summary, GDAS has the following benefits:

1. Compared to previous RL-based and EA-based methods, GDAS makes the searching procedure differentiable, which allows us to end-to-end learn a robust searching rule by gradient descent.
For RL-based and EA-based methods, feedback (reward) is obtained after a prolonged training trajectory, while feedback (loss) in our gradient-based method is instant and is given in every iteration. As a result, the optimization of GDAS is potentially more efficient.
%

2. Instead of using the whole DAG, GDAS samples one sub-graph at one training iteration, accelerating the searching procedure.
Besides, the sampling in GDAS is learnable and contributes to finding a better cell.

3. GDAS delivers a strong empirical performances while using fewer GPU resources. On CIFAR-10, GDAS can finish one searching procedure in several GPU hours and discover a robust neural network with a test error of 2.82\%. On PTB, GDAS discovers a RNN model with a test perplexity of 57.5. Moreover, the networks discovered on CIFAR and PTB can be successfully transferred to ImageNet and WT2.

\section{Related Work}

Recently, researchers have made significant progress in automatically discovering good architectures~\cite{zoph2017NAS,Zoph_2018_CVPR,liu2018hierarchical,Liu_2018_ECCV,real2017large}. Most NAS approaches can be categorized in two modalities: macro search and micro search.

\textbf{Macro search} algorithms aim to directly discover the entire neural networks~\cite{cai2018efficient,brock2018smash,Véniat_2018_CVPR,zoph2017NAS,li2019partial}.
To search convolutional neural networks (CNNs)~\cite{krizhevsky2012imagenet}, typical approaches apply RL to optimize the searching policy to discover architectures~\cite{baker2017designing,cai2018efficient,zoph2017NAS,pmlr-v80-pham18a}.
Baker~et~al.~\cite{baker2017designing} trained a learning agent by Q-learning to sequentially choose CNN layers.
Zoph and Le~\cite{zoph2017NAS} utilized long short-term memory (LSTM)~\cite{hochreiter1997long} as a controller to configure each convolutional layer, such as the filter shape and the number of filters.
In these macro search algorithms~\cite{baker2017designing,cai2018efficient,zoph2017NAS}, the number of possible networks is exponential to the depth of a network, e.g., a depth of 12 can result in more than 10$^{29}$ possible networks~\cite{pmlr-v80-pham18a}.
It is difficult and ineffective to search networks in such a large search space, and, therefore, these macro search methods~\cite{pmlr-v80-pham18a,zoph2017NAS,cai2018efficient} usually limit the CNN models to be shallow, e.g., a depth is less than 12.
Since macro-discovered networks are shallower than deep CNNs~\cite{he2016deep,huang2017densely}, their accuracies are limited.
In contrast, our GDAS allows the network to be much deeper by stacking tens of discovered cells~\cite{Zoph_2018_CVPR} and thus can achieve a better accuracy.
\textbf{Micro search} algorithms aim to discover neural cells and design a neural architecture by stacking many copies of the discovered cells~\cite{Zoph_2018_CVPR,real2019regularized,real2017large,pmlr-v80-pham18a}.
A typical micro search approach is NASNet~\cite{Zoph_2018_CVPR}, which extends the approach of~\cite{zoph2017NAS} to search neural cells in the proposed ``NASNet search space''.
Following NASNet~\cite{Zoph_2018_CVPR}, many researchers propose their methods based on the NASNet search space~\cite{Liu_2018_ECCV,liu2019darts,cai2018efficient,real2019regularized}.
For example, Real~et~al.~\cite{real2019regularized} applied EA algorithm with a simple regularization technique to search neural cells.
Liu~et~al.~\cite{Liu_2018_ECCV} proposed a progressive approach to search cells from shallow to deep gradually.
These micro search algorithms usually take more than 100 GPU days~\cite{Liu_2018_ECCV,Zoph_2018_CVPR}.
Even though some of them reduce the searching cost, they still take more than one GPU day~\cite{liu2019darts}.
Our GDAS is a also micro search algorithm, focusing on search cost reduction. In experiments, we can find a robust network within fewer GPU hours, which is 1000$\times$ less than the standard NAS approach~\cite{Zoph_2018_CVPR}.

\textbf{Improving Efficiency}.
Since NAS algorithms usually require expensive computational resources~\cite{zoph2017NAS,Zoph_2018_CVPR,real2019regularized}, an increasing number of researchers focus on improving the architecture search speed~\cite{cai2018efficient,Liu_2018_ECCV,pmlr-v80-pham18a,Véniat_2018_CVPR,liu2019darts}.
A variety of techniques have been proposed, such as progressive-complexity search stages~\cite{Liu_2018_ECCV}, accuracy prediction~\cite{baker2018accelerating}, HyperNet~\cite{brock2018smash}, Net2Net transformation~\cite{cai2018efficient}, and parameter sharing~\cite{pmlr-v80-pham18a}.
For instance, Cai~et~al.~\cite{cai2018efficient} reused weights of previously discovered networks to amortize the training cost.
Pham~et~al.~\cite{pmlr-v80-pham18a} shared parameters between different child networks to improve the efficiency of the searching procedure.
Brock~et~al.~\cite{brock2018smash} utilized a network to generate model parameters given a discovered network, avoiding fully training from scratch.
Liu~et~al.~\cite{liu2019darts} relaxed the search space to be continuous, so that they can use gradient descent to effectively search cells.
Though these approaches successfully accelerate the architecture search procedure, several GPU days are still required~\cite{Liu_2018_ECCV,cai2018efficient}.
Our GDAS samples individual architecture in a differentiable way to effectively discover architecture. As a result, GDAS can finish the search procedure in several GPU hours on CIFAR-10, which is much faster than these efficient methods.

Contemporary to this work, Xie~et~al.~\cite{xie2018snas} applied a similar technique to relax the discrete candidate sampling as ours. They focus on fixing the inconsistency between the loss of attention-based NAS~\cite{liu2019darts} and their objective. In contrast, we focus on making the sampling procedure differentiable and accelerating the searching procedure.

\section{Methodology}


\subsection{Search Space as a DAG}\label{sec:method-space}

We search for the neural cell in the search space and stack this cell in series to compose the whole neural network.
For CNN, a cell is a fully convolutional network that takes output tensors of previous cells as inputs and generates another feature tensor.
For recurrent neural network (RNN), a cell takes the feature vector of the current step and the hidden state of the previous step as inputs, and generates the current hidden state. For simplification, we take CNN as an example for the following description.

We represent the cell in CNN as a DAG $\gG$ consisting of an ordered sequence of $B$ computational nodes.
Each computational node represents one feature tensor, which is transformed from two previous feature tensors.
This procedure can be formulated as shown in \Eqref{eq:cell} following~\cite{Zoph_2018_CVPR}.
\begin{align}\label{eq:cell}
    \mI_{i} = f_{i,j}( \mI_{j} ) + f_{i,k}( \mI_{k} ) \hspace{2mm}~\mathrm{s.t.~}~\hspace{2mm}j<i\hspace{2mm}\&\hspace{2mm}k<i ,
\end{align}
where $\mI_{i}$, $\mI_{j}$, and $\mI_{k}$ indicate the $i$-th, $j$-th, and $k$-th nodes, respectively. $f_{i,j}$ and $f_{i,k}$ indicate two functions from the candidate function set $\sF$.
We denote the computational nodes of a cell as $B$.
Taking $B=4$ as an example, a cell contains 7 nodes in total, i.e., $\{\mI_{i} | 1 \leq i \leq 7\}$.
$\mI_{1}$ and $\mI_{2}$ nodes are the cell outputs in the previous two layers.
$\mI_{3}$, $\mI_{4}$, $\mI_{5}$, and $\mI_{6}$ nodes are the computational nodes calculated by \Eqref{eq:cell}.
$\mI_{7}$ indicates the output tensor of this cell, which is the concatenation of the four computational nodes, i.e., $\mI_{7} = \mI_{3}^\frown\mI_{4}^\frown\mI_{5}^\frown\mI_{6}$.
In GDAS, the candidate function set $\sF$ contains the following 8 functions:
(1) identity, (2) zeroize, (3) 3x3 depth-wise separate conv, (4) 3x3 dilated depth-wise separate conv, (5) 5x5 depth-wise separate conv, (6) 5x5 dilated depth-wise separate conv, (7) 3x3 average pooling, (8) 3x3 max pooling.
We use the same candidate function set $\sF$ as~\cite{liu2019darts}, which is similar to \cite{Zoph_2018_CVPR} but removes some unused functions and adds some useful functions.

\begin{figure}[t!]
\begin{center}
\includegraphics[width=\linewidth]{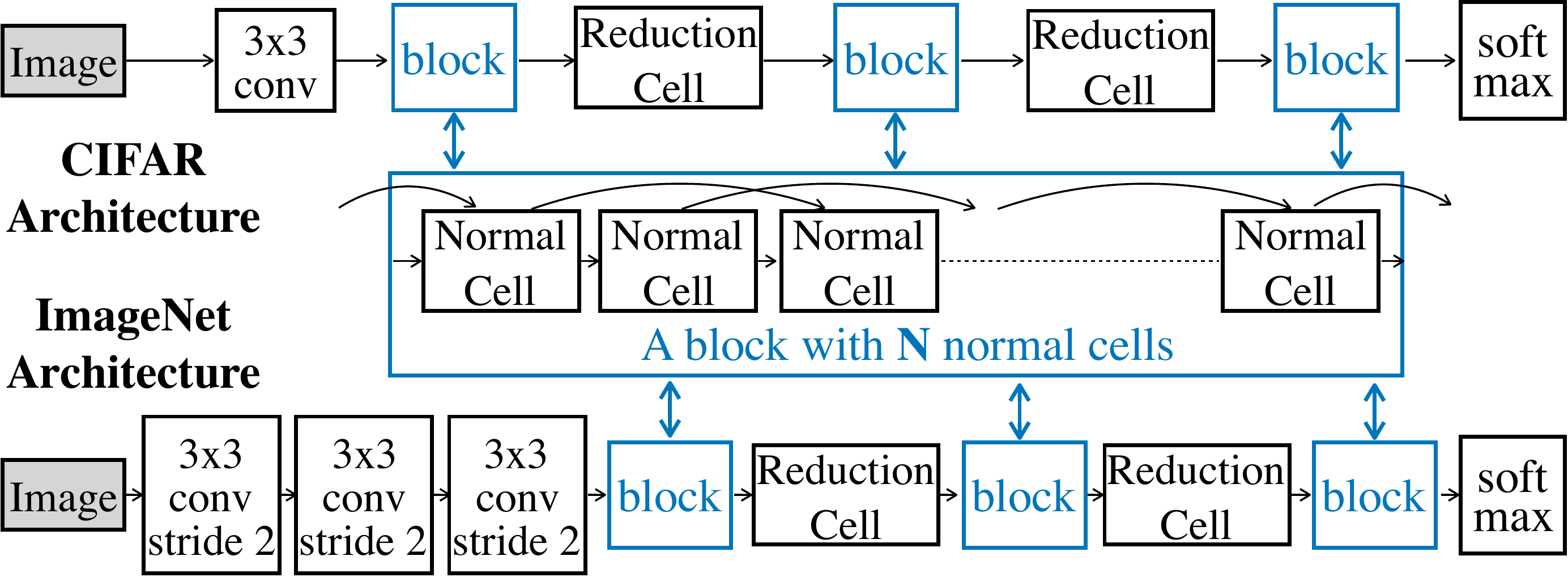}
\end{center}
\caption[Captioning]{
The strategy to design CIFAR architecture (top) and ImageNet architecture (bottom) based on the discovered cell.
We use the same block structure (middle) in two cases.
Both normal and reduction cells receive the outputs of two previous cells as inputs, as illustrated in the middle.
}
\label{fig:cell2net}
\end{figure}

\textbf{From cell to network}.
We search for two kinds of cells, i.e., a normal cell and a reduction cell.
For the normal cell, each function in $\sF$ has the stride of 1.
For the reduction cell, each function in $\sF$ has the stride of 2.
Once we discover one normal cell and one reduction cell, we stack many copies of these discovered cells to make up a neural network.
As shown in \Figref{fig:cell2net}, for the CIFAR architecture, we stack $N$ normal cells as one block.
Given an image, it first forwards through the network head part, i.e., one 3 by 3 convolutional layer.
It then forwards through three blocks with two reduction cells in between. The ImageNet architecture is similar to the CIFAR architecture, but the network head part consists of three 3 by 3 convolutional layers.
We follow~\cite{liu2019darts} to setup these two overall structures.

\subsection{Searching by Differentiable Model Sampling}\label{sec:method-sample}

Formally, we denote a neural architecture as $\alpha$ and the weights of this neural architecture as $\omega_{\alpha}$.
The goal of NAS is to find an architecture $\alpha$, which can achieve the minimum validation loss after being trained by minimizeing the training loss, as shown in \Eqref{eq:objective}.
\begin{align}\label{eq:objective}
    \min_{\alpha} \sE_{(x',y')\sim{\sD}_{V}} -\log\mathrm{Pr}(y'|x'; \alpha, \omega_{\alpha}^{*}) ,  \nonumber\\
    \mathrm{s.t.}\hspace{1mm}\omega_{\alpha}^{*} = \arg{\min}_{\omega} \sE_{(x,y)\sim{\sD}_{T}} -\log\mathrm{Pr}(y|x; \alpha, \omega_{\alpha}) ,
\end{align}
\noindent where $\omega_{\alpha}^{*}$ is the best weight of $\alpha$ and achieves the minimum training loss.
We use the negative log likelihood as the training objective, i.e., $-\log\mathrm{Pr}$.
${\sD}_{T}$ and ${\sD}_{V}$ indicate the training set and the validation set, respectively.
$(x,y)$ and $(x',y')$ are the data associated with its label, which are sampled from ${\sD}_{T}$ and ${\sD}_{V}$, respectively.

An architecture $\alpha$ consists of many copies of the neural cell.
This cell is sampled from the search space represented by $\gG$.
Specifically, between node$_{i}$ and node$_{j}$, we sample one transformation function from $\sF$ from a discrete probability distribution $\gT_{i,j}$.
During the search, we calculate each node in a cell as:
\begin{align}\label{eq:sample}
    \mI_{i} = \sum_{j=1}^{i-1} ~f_{i,j}~ ( \mI_{j}; \mW_{f_{i,j}} )  \hspace{0.3cm} \mathrm{s.t.} \hspace{0.3cm} f_{i,j} \sim \gT_{i,j},
\end{align}
\noindent where $f_{i,j}$ is sampled from $\gT_{i,j}$ and $\mW_{f_{i,j}}$ is its associated weight.
The discrete probability distribution $\gT_{i,j}$ is characterized by a learnable probability mass function as in \Eqref{eq:softmax}:
\begin{align}\label{eq:softmax}
    \mathrm{Pr}(f_{i,j}=\sF_{k}) = \frac{\exp(\mA_{i,j}^{k})}{\sum_{k'=1}^{K} \exp(\mA_{i,j}^{k'})} ,
\end{align}
\noindent where $\mA_{i,j}^{k}$ is the $k$-th element of a $K$-dimensional learnable vector $\mA_{i,j} \in \sR^{K}$, and $\sF_{k}$ indicates the $k$-th function in $\sF$.
$K$ is the cardinality of $\sF$, i.e., $K=|\sF|$.
Actually, $\mA_{i,j}$ encodes the sampling distribution of the function between node$_{i}$ and node$_{j}$.
As a result, the sampling distribution of a neural cell is encoded by all $\mA_{i,j}$, i.e., $\gA = \{\mA_{i,j}\}$.

Given \Eqref{eq:sample} and \Eqref{eq:softmax}, we can obtain $\alpha$ and $\omega$, and thus can calculate $\mathrm{Pr}(y|x; \alpha, \omega)$ in \Eqref{eq:objective}.
However, since \Eqref{eq:sample} needs to sample from a discrete probability distribution, we cannot back-propagate gradients through $\mA_{i,j}$ in \Eqref{eq:softmax} to optimize $\mA_{i,j}$.
To allow back-propagation, we first use the Gumbel-Max trick~\cite{gumbel1954statistical,maddison2014sampling} to re-formulate \Eqref{eq:sample} as \Eqref{eq:Gumbel-max}, which provides an efficient way to draw samples from a discrete probability distribution.
\begin{align}
  \mI_{i}      ~&~ = \sum_{j=1}^{i-1} \sum_{k=1}^{K} {\vh_{i,j}^{k}}~\sF_{k} ( \mI_{j}; \mW_{i,j}^{k} ) , \label{eq:Gumbel-max}\\
  \mathrm{s.t.}~&~ {\vh_{i,j}} = \mathrm{one\_hot}(\arg\max_{k}( \mA_{i,j}^{k} + \vo_{k} ) ), \label{eq:one-hot}
\end{align}
\noindent where $\vo_{k}$ are i.i.d samples drawn from Gumbel~(0,1)\footnote{$\vo_{i}=-\log(-\log(u))$ with $u\sim\mathrm{Unif~[0,1]}$}, and
$\vh_{i,j}^{k}$ is the $k$-th element of $\vh_{i,j}$.
$\mW_{i,j}^{k}$ is the weight of $\sF_{k}$ for the transformation function between node$_i$ and node$_j$.
Then, we use the softmax function to relax the $\arg\max$ function so as to make \Eqref{eq:Gumbel-max} being differentiable~\cite{jang2017categorical,maddison2017concrete}.
Formally, we use \Eqref{eq:gumbel-soft} to approximate \Eqref{eq:Gumbel-max}.
\begin{align}\label{eq:gumbel-soft}
  \tilde{\vh}_{i,j}^{k} = \frac{\exp((\log(\mathrm{Pr}(f_{i,j}=\sF_{k})) + \vo_{k}) / \tau)}{ \sum_{k'=1}^{K} \exp( (\log(\mathrm{Pr}(f_{i,j}=\sF_{k'})) + \vo_{k'}) /\tau ) } ,
\end{align}
\noindent where $\tau$ is the softmax temperature.
When $\tau \rightarrow 0$, $\tilde{\vh}_{i,j}^{k} = {\vh}_{i,j}^{k}$.
When $\tau \rightarrow \infty$, each element in $\tilde{\vh}$ will be the same and the approximated distribution will be smooth.
To be noticed, we use the $\arg\max$ function in \Eqref{eq:Gumbel-max} during the forward pass but the $\mathrm{soft}\max$ function in \Eqref{eq:gumbel-soft} during the backward pass to allow gradient back-propagation.

\begin{algorithm}[t]
\small
\caption{The GDAS algorithm}
\label{alg:Train}

  \begin{algorithmic}
    \Require split the training set into two disjoint sets: $\sD_{T}$ and $\sD_{V}$; randomly initialized $\gA$ and $\gW$, and the batch size $n$
    \While{not converge} \Comment{searching an architecture}
    	\State Sample batch of data {$\sD_{t} = \{ (x_{i},y_{i})\}_{i=1}^{n}$ from $\sD_{T}$}
    	\State Calculate $L_{T}=\sum_{i=1}^{n} \ell(x_{i},y_{i})$ based on \Eqref{eq:loss}
    	\State Update $\gW$ by gradient descent: $\gW=\gW - \triangledown_{\gW} L_{T}$
    	\State Sample batch of data {$\sD_{v} = \{ (x_{i},y_{i})\}_{i=1}^{n}$ from $\sD_{V}$}
    	\State Calculate $L_{V}=\sum_{i=1}^{n} \ell(x_{i},y_{i})$ based on \Eqref{eq:loss}
    	\State Update $\gA$ by gradient descent: $\gA=\gA - \triangledown_{\gA} L_{V}$
    \EndWhile
    \State Derive the final architecture from $\gA$
    \State Optimize the architecture on the training set
  \end{algorithmic}
  
\end{algorithm}

{\bf Training.}
Reviewing the objective of NAS in \Eqref{eq:objective}, the main challenge is learning to find architecture $\alpha$.
By utilizing \Eqref{eq:gumbel-soft}, we can make the sampling procedure differentiable and learn a distribution of neural cells (representing architectures).
However, it is still intractable to directly solve \Eqref{eq:objective}, because the nested formulation in \Eqref{eq:objective} needs to calculate high order derivatives.
In practice, to avoid calculating high order derivatives, we apply the alternative optimization strategy to update the sampling distribution $\gT_{\gA}$ and the weights of all functions $\gW$ in an iterative way.
Given one data sample $x$ and its associated label $y$, we calculate the loss as:
\begin{align}\label{eq:loss}
  \ell (x,y)    ~&~= -\log\mathrm{Pr}(y|x; \alpha, \omega_{\alpha}), \\
  \mathrm{s.t.} ~&~ \alpha \sim \gT_{\gA} \hspace{2mm}\&\hspace{2mm}\omega_{\alpha} \subset \gW ,
\end{align}
\noindent where $\gT_{\gA}$ is the distribution encoded by ${\gA}$ and $\gW = \{\mW_{i,j}^{k}\}$ represents the weights of all functions in all cells of the network.
Note that, for one data sample, it first samples $\alpha$ from $\gT_{\gA}$ and then calculates the network output only on its associated weight $\omega_{\alpha}$, which is a part of $\gW$.
As shown in \Algref{alg:Train}, we apply the alternative optimization strategy (AOS) to update $\gA$ based on the validation losses from $\sD_{V}$ and update $\gW$ based on the training losses from $\sD_{T}$.
It is essential to train $\gW$ on $\sD_{T}$ and $\gA$ on $\sD_{V}$, because (1) this strategy is theoretically sound with the objective \Eqref{eq:objective}; and (2) this strategy can improve the generalization ability of the searched structure.

{\bf Architecture}.
After training, we need to derive the final architecture from the learned $\gA$.
Each node$_{i}$ connects with $T$ previous nodes.
Following the previous works, we use $T=2$ for CNN~\cite{Zoph_2018_CVPR,liu2019darts} and $T=1$ for RNN~\cite{pmlr-v80-pham18a,liu2019darts}.
Suppose $\Omega$ is the candidate index set, we derive the final architecture by the following procedure:
(1) define the importance of the connection between node$_{i}$ and node$_{j}$ as: $\max_{k\in\Omega} \mathrm{\Pr}(f_{i,j}=\sF_{k})$.
(2) for each node$_{i}$, retain $T$ connections with the maximum importance from the previous nodes.
(3) for the retained connection between node$_{i}$ and node$_{j}$, we use the function $\sF_{\arg\max_{k\in\Omega} \mathrm{\Pr}(f_{i,j}=\sF_{k})}$.
$\Omega$ is $\{1,...,K\}$ by default.

{\bf Acceleration.}
In \Eqref{eq:Gumbel-max}, $\vh_{i,j}$ is a one-hot vector. As a result, in the forward procedure, we only need to calculate the function $\sF_{\arg\max(\vh_{i,j})}$.
During the backward procedure, we only back-propagate the gradient generated at the $\arg\max(\tilde{\vh}_{i,j})$.
In this way, we can save most computation time and also reduce the GPU memory cost by about $|\sF|$ times.
Within one training batch, a different cell will produces a different $\vh_{i,j}$, which was shared for each training examples.

One benefit of this acceleration trick is that it allows us to directly search on the large-scale dataset (e.g., ImageNet) due to the saved GPU memory. We did some experiments to directly search on ImageNet using the same hyper-parameters as on the small datasets, however, failed to obtain a good performance.
Searching on a large-scale dataset might require different hyper-parameters and needs careful tuning. We will explore this in our future work.

\begin{table*}[t!]
\centering
\setlength{\tabcolsep}{5.3pt}
\begin{tabular}{| c | l | c | c | c | c | c | c |} \hline\hline

  \multirow{2}{*}{\textbf{Type}} &  \multirow{2}{*}{\textbf{Method}} & \multirow{2}{*}{\textbf{Venue}} & \multirow{2}{*}{\textbf{GPUs}} & \textbf{Times} & \textbf{Params} & \textbf{Error} on & \textbf{Error} on \\
    &  & & & (days) & (million) & CIFAR-10 & CIFAR-100  \\ \hline
 \multirow{2}{*}{\makecell{Human\\expert}}
    &  ResNet + CutOut~\cite{he2016deep}                & CVPR16  & $-$ & $-$   & 1.7  & 4.61          & 22.10 \\
    &   DenseNet-BC~\cite{huang2017densely}             & CVPR17  & $-$ & $-$   & 25.6 & 3.46          & 17.18 \\
      \hline\hline
\multirow{6}{*}{\makecell{Macro\\search\\space}}
    &  MetaQNN~\cite{baker2017designing}                & ICLR17  & 10  & 8-10  & 11.2 & 6.92          & 27.14 \\
    &  Net Transformation~\cite{cai2018efficient}       & AAAI18  & 5   & 2     & 19.7 & 5.70          &  $-$  \\
    &  SMASH~\cite{brock2018smash}                      & ICLR18  & 1   & 1.5   & 16.0 & 4.03          &  $-$  \\
    &  NAS~\cite{zoph2017NAS}                           & ICLR17  & 800 & 21-28 & 7.1  & 4.47          &  $-$  \\
    &  NAS + more filters~\cite{zoph2017NAS}            & ICLR17  & 800 & 21-28 & 37.4 & 3.65          &  $-$  \\
    &  ENAS~\cite{pmlr-v80-pham18a}                     & ICML18  & 1   & 0.32  & 38.0 & 3.87          &  $-$  \\
      \hline\hline
\multirow{16}{*}{\makecell{Micro\\search\\space}}
    &  Hierarchical NAS~\cite{liu2018hierarchical}      & ICLR18  & 200  & 1.5         & 61.3 & 3.63          &  $-$  \\
    &  Progressive NAS~\cite{Liu_2018_ECCV}             & ECCV18  & 100  & 1.5         & 3.2  & 3.63          &  19.53  \\
    &  NASNet-A~\cite{Zoph_2018_CVPR}                   & CVPR18  & 450  & 3-4         & 3.3  & 3.41          & $-$ \\
    &  NASNet-A + CutOut~\cite{Zoph_2018_CVPR}          & CVPR18  & 450  & 3-4         & 3.3  & \textbf{2.65} & $-$ \\
    &  ENAS~\cite{pmlr-v80-pham18a}                     & ICML18  & 1    & 0.45        & 4.6  & 3.54          &  19.43 \\
    &  ENAS + CutOut~\cite{pmlr-v80-pham18a}            & ICML18  & 1    & 0.45        & 4.6  & 2.89          &  $-$ \\
    &  DARTS (1st) + CutOut~\cite{liu2019darts}   & ICLR19  & 1    & 0.38   & 3.3   & 3.00          & $-$ \\
    &  DARTS (2nd) + CutOut~\cite{liu2019darts}   & ICLR19  & 1    & 1           & 3.4  & 2.82          & \textbf{17.54}$\dagger$ \\
    &  GHN + CutOut~\cite{zhang2019graph}               & ICLR19  & 1    & 0.84        & 5.7  & 2.84          & $-$ \\
    &  NAONet~\cite{luo2018neural}                      & NeurIPS18  & 200  & 1           & 10.6 & 3.18          & $-$ \\
    &  AmoebaNet-A + CutOut~\cite{real2019regularized}  & AAAI19 & 450  & 7           & 3.1  & 3.12          & 18.93$\dagger$ \\
    \cline{2-8}
    & GDAS [C=36,N=6]                                   & CVPR19 & 1    & 0.21        & 3.4  & 3.87          & 19.68 \\
    & GDAS [C=36,N=6] + CutOut                          & CVPR19 & 1    & 0.21        & 3.4  & 2.93          & 18.38 \\
    & GDAS (FRC) [C=36,N=6]                             & CVPR19 & \textbf{1} & \textbf{0.17} & \textbf{2.5}  &  3.75   & 19.09 \\
    & GDAS (FRC) [C=36,N=6] + CutOut                    & CVPR19 & \textbf{1} & \textbf{0.17} & \textbf{2.5}  &  2.82   & 18.13 \\
    \hline
    \hline
\end{tabular}
\vspace{2mm}
\caption{
Classification errors of GDAS and baselines on CIFAR.
$\dagger$ indicates the results trained using our setup.
``FRC'' indicates that we fix the reduction cell and only search the normal cell.
Note that researchers might run their algorithms on different kinds of machines.
The searching costs are derived from the original papers, and we did not normalize them across different GPUs. Our experiments are based on the V100 GPU; and if we run on Titan 1080Ti, the searching cost will increase to about seven GPU hours.
}
\vspace{-2mm}
\label{table:CIFAR}
\end{table*}

\subsection{Discussion on the Reduction Cell}\label{sec:method-reduction}

\begin{figure}[t!]
\begin{center}
\includegraphics[width=\linewidth]{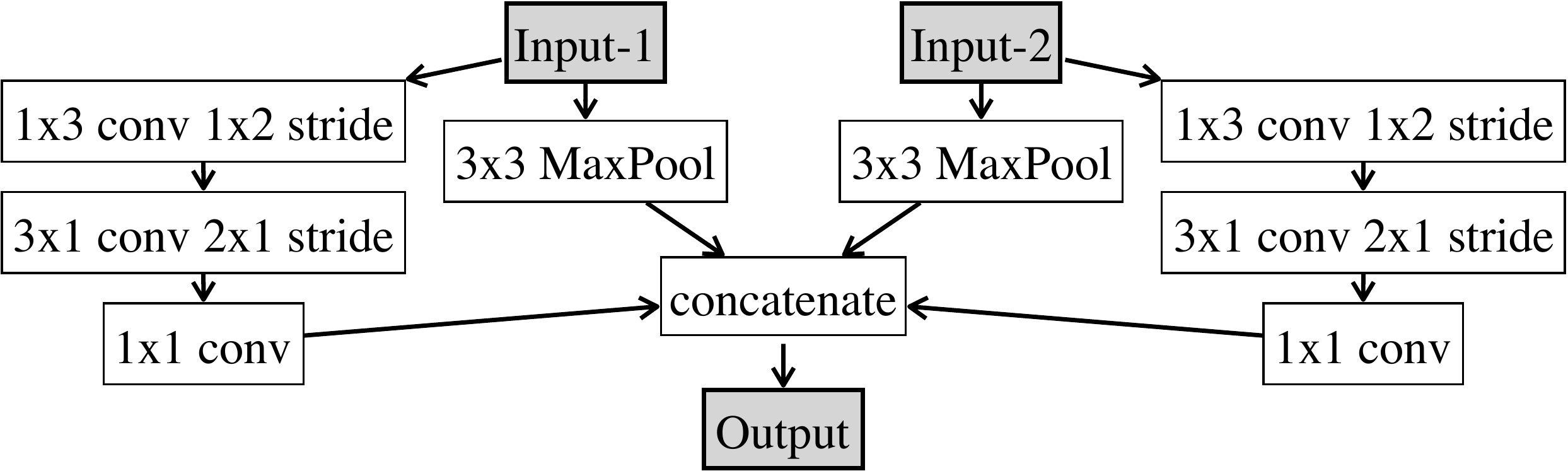}
\end{center}
\caption[Captioning]{
The designed reduction cell.
``1x3 conv 1x2 stride'' indicates a convolutional layer with 1 by 3 kernel and 1 by 2 stride.
}
\vspace{-2mm}
\label{fig:reduction}
\end{figure}

Revisiting state-of-the-art architectures designed by human experts, AlexNet~\cite{krizhevsky2012imagenet} and VGGNet~\cite{simonyan2015very} use the max pooling to reduce the spatial dimension; ResNet~\cite{he2016deep} uses a convolutional layer with stride of 2; and DenseNet~\cite{huang2017densely} uses a 1 by 1 convolutional layer followed by average pooling to reduce dimension.
These human-designed reduction cells are simple and effective.
The automatically discovered reduction cells are also usually similar and simple~\cite{Zoph_2018_CVPR,pmlr-v80-pham18a,liu2019darts}. For example, the reduction cell discovered by~\cite{liu2019darts} only has max pooling and identity operations.

Most human-designed and automatically discovered reduction cells are simple and can achieve a high accuracy.
Moreover, compared to searching one normal cell, jointly searching a normal cell and a reduction cell will greatly increase the search space and make the optimization difficult.
We hope to find a better network by fixing the reduction cell.
Inspired by~\cite{simonyan2015very,liu2019darts}, we design a fixed reduction cell as shown in \Figref{fig:reduction}.
In the experiments, with this human-designed reduction cell, GDAS finds a better architecture, yielding fewer parameters and higher accuracy.

\section{Experimental Study}


\subsection{Datasets}\label{sec:exp-data}

\textbf{CIFAR-10} and \textbf{CIFAR-100}~\cite{krizhevsky2009learning} consist of 50K training images and 10K test images.
CIFAR-10 categorizes images into 10 classes, while CIFAR-100 has 100 classes.

\textbf{ImageNet}~\cite{russakovsky2015imagenet} is a large-scale and well-known benchmark for image classification. It contains 1K classes, 1.28 million images for training, and 50K images for validation.

\textbf{Penn Treebank (PTB)}~\cite{marcus1993building} is a corpus consisting of over 4.5 million words of American English words. We pre-process PTB following~\cite{merity2017pointer,merity2018regularizing}.

\textbf{WikiText-2 (WT2)}~\cite{merity2017pointer} is a collection of 2 million tokens from the set of verified Good and Featured articles on Wikipedia. The training set contains 600 articles with 2,088,628 tokens. The validation set contains 60 articles with 217,646 tokens. The test set contains 60 articles with 245,569 tokens.

\subsection{Search for CNN}\label{sec:exp-cnn}

\textbf{CNN Searching Setup}.
The neural cells for CNN are searched on CIFAR-10 following~\cite{zoph2017NAS,Zoph_2018_CVPR,Liu_2018_ECCV,pmlr-v80-pham18a}.
We randomly split the official training images into two groups, with each group containing 25K images. One group is used as the training set $\sD_{T}$ in \Algref{alg:Train}, and the other is used as the validation set $\sD_{V}$ in \Algref{alg:Train}.
The candidate function set $\sF$ has 8 different functions as introduced in \Secref{sec:method-space}.
The default hyper-parameters for each function in $\sF$ are the same in~\cite{liu2019darts,Zoph_2018_CVPR,Liu_2018_ECCV}.
By default, we set the number of initial channels in the first convolution layer $C$ as 16; set the number of computational nodes in a cell $B$ as 4; and the number of layers in one block $N$ as 2.
We train the model by 240 epochs in total.
For $\omega$, we use the SGD optimization. We start with a learning rate of 0.025 and anneal it down to 1e-3 following a cosine schedule. We use the momentum of 0.9 and the weight decay of 3e-4.
For $\alpha$, we use the Adam optimization~\cite{kingma2015adam} with the learning rate of 3e-4 and the weight decay of 1e-3.
The $\tau$ is initialized as 10 and is linearly reduced to {0.1}.
To search the normal cell and the reduction cell on CIFAR-10, our GDAS takes about five hours to finish the search procedure on a single NVIDIA Tesla V100 GPU.
As discussed in \Secref{sec:method-reduction}, we also run experiments to only search the normal cell and fix the reduction cell as shown in~\Figref{fig:reduction}, denoted as GDAS (FRC).
When we use GDAS (FRC) to search on CIFAR-10, it takes less than four hours to finish one search procedure.
\textit{Following~\cite{liu2019darts}, we run GDAS 4 times with different random seeds and pick the best cell based on its validation performance. This procedure can reduce the high variance of the searched results, especially when searching the RNN structure.}

\textbf{Clarifications on the searching cost (GPU days) of different methods.}
The searching costs listed in \Tabref{table:CIFAR} and \Tabref{table:ImageNet} are \textbf{not normalized} across different GPU devices. Different algorithms might run on different machines, and we simply refer the searching costs reported in their papers.\footnote{It is difficult for us to run all algorithms on the same GPU.}
If we use other GPU devices, the searching cost of ``GDAS (FRC)'' could be a different number. For example, if we use Titan 1080Ti, the search cost will increase to about seven GPU hours.

\textbf{Discussion on the acceleration step}. If we do not apply the acceleration step introduced in \Secref{sec:method-sample}, each iteration will cost $|\sF|\text{=}8\times$ more time and GPU memory than GDAS. In the same time, without this acceleration step, it requires less training epochs to converge but still costs more time than applying the acceleration step.

\begin{table*}[t!]
\centering
\setlength{\tabcolsep}{4.5pt}
\begin{tabular}{| c | l | c | c | c | c | c | c | c |} \hline\hline

  \multirow{2}{*}{\textbf{Type}} &  \multirow{2}{*}{\textbf{Method}} & \multirow{2}{*}{\textbf{Venue}} & \multirow{2}{*}{\textbf{GPUs}} & \textbf{Times} & \multicolumn{2}{c|}{Test Error (\%)} & \textbf{Params} & $+\times$  \\ \cline{6-7}
    &  & & & (days) & Top-1 & Top-5 & (million) & (million) \\ \hline

 \multirow{3}{*}{\makecell{Human expert}}
    &  Inception-v1~\cite{szegedy2015going}             & CVPR15  & $-$  & $-$   & 30.2 & 10.1    &   6.6  &   1448  \\
    &  MobileNet-V2~\cite{sandler2018mobilenetv2}       & CVPR18  & $-$  & $-$   & 28.0 &  $-$    &   3.4  &   300   \\
    &  ShuffleNet~\cite{Zhang_2018_CVPR}                & CVPR18  & $-$  & $-$   & 26.3 &  $-$    &$\sim$5 &   524   \\
      \hline\hline                                      
\multirow{12}{*}{\makecell{Micro search space}}
    &  Progressive NAS~\cite{Liu_2018_ECCV}             & ECCV18  & 100  & 1.5   & {25.8}  & 8.1  &   5.1  & 588   \\
    &  NASNet-A~\cite{Zoph_2018_CVPR}                   & CVPR18  & 450  & 3-4   & 26.0  & 8.4    &   5.3  & 564   \\
    &  NASNet-B~\cite{Zoph_2018_CVPR}                   & CVPR18  & 450  & 3-4   & 27.2  & 8.7    &   5.3  & 488   \\
    &  NASNet-C~\cite{Zoph_2018_CVPR}                   & CVPR18  & 450  & 3-4   & 27.5  & 9.0    &   4.9  & 558   \\
    &  DARTS (2nd)~\cite{liu2019darts}                  & ICLR19  & 1    & 1     & 26.9  & 9.0    &   4.9  & 595   \\
    &  GHN~\cite{zhang2019graph}                        & ICLR19  & 1    & 0.84  & 27.0  & 8.7    &   6.1  & 569   \\
    &  AmoebaNet-A~\cite{real2019regularized}           & AAAI19  & 450  & 7     & 25.5  & 8.0    &   5.1  & 555 \\
    &  AmoebaNet-B~\cite{real2019regularized}           & AAAI19  & 450  & 7     & 26.0  & 8.5    &   5.3  & 555   \\
    &  AmoebaNet-C~\cite{real2019regularized}           & AAAI19  & 450  & 7     & \textbf{24.3}  & \textbf{7.6}    &   6.4  & 570   \\
    \cline{2-9}
    & GDAS [C=50,N=4]                                   & CVPR19  & 1    & 0.21  & 26.0  & 8.5    &   5.3  & 581   \\
    & GDAS (FRC) [C=52,N=4]                             & CVPR19  & \textbf{1}    & \textbf{0.17} & 27.5 & 9.1 & \textbf{4.4}  & \textbf{497} \\
    \hline\hline
\end{tabular}
\vspace{2mm}
\caption{
Top-1 and top-5 errors of GDAS and baselines on ImageNet. $+\times$ indicates the number of multiply-add operations. We refer results reported in~\cite{liu2019darts} for Progressive NAS, NASNet, and AmoebaNet.
}
\vspace{-2mm}
\label{table:ImageNet}
\end{table*}

\textbf{Results on CIFAR}.
After the searching procedure, we use C=36, B=4, and N=6 to form a CNN.
Following the previous works~\cite{liu2019darts,pmlr-v80-pham18a,Zoph_2018_CVPR}, we train the network by 600 epochs in total. We start the learning rate of 0.025 and reduce it to 0 with the cosine learning rate scheduler.
We set the probability of path dropout as 0.2 and the auxiliary tower with the weight of 0.4~\cite{zoph2017NAS}.
We use the standard pre-processing and data augmentation, i.e., randomly cropping, horizontally flipping, normalization, and CutOut~\cite{CUTOUT,zhong2017random}.

We compare the models discovered by our approach with other state-of-the-art models in \Tabref{table:CIFAR}.
The models discovered by the macro search algorithms obtain a higher error than the models discovered by the micro search algorithms.
Using GDAS, we discover a model with 3.3M parameters, which achieves 2.93\% error on CIFAR-10.
Using GDAS (FRC), we discover a model with only 2.5M parameters, which achieves 2.82\% error on CIFAR-10.
NASNet-A achieves a lower error rate than ours, but it contains more than 80\% of the parameters than the model discovered by GDAS (FRC).
Notably, our GDAS discovers a comparable model with the state-of-the-art, whereas the searching cost of our approach is much less than the others.
For example, GDAS (FRC) takes less than 4 hours on a single V100 GPU, which is about 0.17 GPU days.
It is faster than NASNet by almost 10$^{4}$ times.
ENAS is a recent work that focuses on accelerating the searching procedure.
ENAS is very efficient, whereas our GDAS (FRC) is three times faster than ENAS.

\textbf{Results on ImageNet}.
Following~\cite{Zoph_2018_CVPR,liu2019darts,pmlr-v80-pham18a,sandler2018mobilenetv2}, we use the ImageNet-mobile setting, in which the input size is 224$\times$224 and the number of multiply-add operations is restricted to be less than 600M.
We train models by SGD with 250 epochs and use the batch size of 128.
We initialize the learning rate of 0.1 and reduce it by 0.97 after each epoch.

\begin{table}[t!]
\centering
\setlength{\tabcolsep}{4pt}
\begin{tabular}{| c | c | c | c | c |} \hline\hline

\multirow{2}{*}{Architecture}          & \multicolumn{2}{c|}{Perplexity} & Params & Search Cost \\\cline{2-3}
                                       &     val     &      test        & (M)    & (GPU days) \\\hline
V-RHN~\cite{zilly2017recurrent}        &   67.9      &   65.4           & 23     &   $-$      \\
LSTM~\cite{merity2018regularizing}     &   60.7      &   58.8           & 24     &   $-$      \\
LSTM + SC~\cite{merity2018regularizing}&   60.9      &   58.3           & 24     &   $-$       \\
LSTM + SE~\cite{yang2018breaking}      &   58.1      &   56.0           & 22     &   $-$       \\
\hline\hline
NAS~\cite{zoph2017NAS}                 &    $-$      &   64.0           & 25     & 10$^{4}$    \\
ENAS~\cite{pmlr-v80-pham18a}           &    60.8     &   58.6           & 24     & 0.5         \\
DARTS (1st)~\cite{liu2019darts}        &    60.2     &   57.6           & 23     & \textbf{0.13} \\
DARTS (2nd)~\cite{liu2019darts}        &\textbf{58.1}&\textbf{55.7}     &\textbf{23}&0.25 \\\hline
GDAS                                   &    59.8     &      57.5        &\textbf{23}&    0.4     \\\hline
\hline
\end{tabular}
\vspace{2mm}
\caption{
Comparison \textit{w.r.t.} the perplexity of different language models on PTB (lower perplexity is better).
V-RHN indicates Variational RHN~\cite{zilly2017recurrent}.
LSTM + SC indicates LSTM with skip connection~\cite{merity2018regularizing}.
LSTM + SE indicates LSTM with 15 softmax experts~\cite{yang2018breaking}.
The first four models are designed by human experts, and the last four models are automatically searched by machine.
}
\vspace{-2mm}
\label{table:PTB}
\end{table}

We compare our results on ImageNet with the other methods in \Tabref{table:ImageNet}.
Most algorithms in \Tabref{table:ImageNet} take more than 1000 GPU days to discover a good CNN cell.
DARTS~\cite{liu2019darts} uses minimum resources among the compared algorithms, whereas ours is even faster than DARTS~\cite{liu2019darts} by more than 10 times.
For GDAS (FRC), we use C=52 and N=4 to construct the model following the setting in~\cite{liu2019darts}.
For GDAS, if we use C=52 and N=4, the number of multiply-add operations will be larger than 600 MB, and thus we use C=50 to restrict it to be less than 600MB.
Our model, GDAS (FRC) [C=52,N=4], costs about 20\% less multiply-add operations than \cite{liu2019darts} but obtains the same top-5 error.
AmoebaNet-A and Progressive NAS achieve a slightly lower test error than ours.
However, their methods cost a prohibitive amount of GPU resources.
The results in \Tabref{table:ImageNet} show the discovered cell on CIFAR-10 can be successfully transferred to ImageNet and achieve competitive performance.

\subsection{Search for RNN}\label{sec:exp-rnn}

\begin{table}[t!]
\centering
\setlength{\tabcolsep}{4pt}
\begin{tabular}{| c | c | c | c | c |} \hline\hline

 \multirow{2}{*}{Architecture}          &  \multicolumn{2}{c|}{Perplexity} & Params & Search Cost \\\cline{2-3}
                                        &     val     &      test        & (M)    & (GPU days)  \\ \hline
LSTM + AL~\cite{inan2017tying}          &  91.5       &   82.0           & 28     &   $-$       \\
LSTM~\cite{merity2018regularizing}      &  69.1       &   65.9           & 33     &   $-$       \\
LSTM + SC~\cite{merity2018regularizing} &  69.1       &   65.9           & 23     &   $-$       \\
LSTM + SE~\cite{yang2018breaking}       &  66.0       &   63.3           & 33     &   $-$       \\
\hline\hline
ENAS~\cite{pmlr-v80-pham18a}            &  72.4       &   70.4           &\textbf{33}& 0.5          \\
DARTS (2nd)~\cite{liu2019darts}         &  71.2       &   69.6           &\textbf{33}&\textbf{0.25} \\\hline
GDAS                                    &\textbf{71.0}&\textbf{69.4}     &\textbf{33}& 0.4 \\\hline
\hline
\end{tabular}
\vspace{2mm}
\caption{
Comparison with different language models on WT2 (lower perplexity is better).
LSTM + AL indicates LSTM with augmented loss~\cite{inan2017tying}. Other notation is the same as in \Tabref{table:PTB}.
}
\vspace{-2mm}
\label{table:WT2}
\end{table}

\textbf{RNN Searching Setup}.
The neural cells for RNN are searched on PTB with the splits following~\cite{liu2019darts,pmlr-v80-pham18a}
The candidate function set $\sF$ contain 5 functions, i.e., zeroize, Tanh, ReLU, sigmoid, and identity.
We use $B\text{=}9$ to search the RNN cell.
The RNN model consists of one word embedding layer with a hidden size of 300, one RNN cell with a hidden size of 300, and one decoder layer.
We train the model by 200 epochs with a batch size of 128 and a BPTT length of 35.
We optimize $\omega$ by Adam with a learning rate of 20 and a weight decay of 5e-7.
We optimize $\alpha$ by Adam with a learning rate of 3e-3 and a weight decay of 1e-3.
Other setups are the same in~\cite{pmlr-v80-pham18a,liu2019darts}.

\textbf{Results on PTB}.
We evaluate the RNN model formed by the discovered recurrent cell on PTB.
We use a batch size of 64 and a hidden size of 850.
We train the model using the A-SGD by 2000 epochs.
The learning rate is fixed as 20 and the weight decay is 8e-7.
DARTS~\cite{liu2019darts} and ENAS~\cite{pmlr-v80-pham18a} greatly reduce the search cost compared to previous methods.
Our GDAS incurs a lower search cost than all the previous methods.
Note that our code is not heavily optimized and the theoretical search cost should be less than the one reported in \Tabref{table:PTB}.

We compare different RNN models in \Tabref{table:PTB}.
The model discovered by GDAS achieves a validation perplexity of 59.8 and a test perplexity of 57.5.
The performance of our discovered RNN is on par with the state-of-the-art models in \Tabref{table:PTB}.
LSTM + SE~\cite{yang2018breaking} obtains better results than ours, but it is an ensemble method using mixture of softmax.
By applying the SE technique~\cite{yang2018breaking}, GDAS can achieve the lower perplexity without doubt.
LSTM~\cite{merity2018regularizing} is an extensively tuned model, whereas our automatically discovered model is superior to it.
Compared to other efficient approaches, the search cost of GDAS is the lowest.

\begin{table}[t!]
\centering
\setlength{\tabcolsep}{4pt}
\begin{tabular}{| c | c | c |} \hline\hline

 Architecture     & Params       & Error on CIFAR-10  \\\hline
 GDAS-N + GDAS-R  & 3.3          &  2.93              \\
 GDAS-N + FIX-R   & 3.0          &  2.87              \\ FRC-N  + GDAS-R  & 2.9          &  2.84              \\
 FRC-N  + FIX-R   & \textbf{2.5} &  \textbf{2.82}     \\
\hline\hline
\end{tabular}
\vspace{2mm}
\caption{
Different normal and reduction cell combinations.
GDAS-N and GDAS-R indicate the normal and reduction cells discovered by GDAS, {respectively}.
FRC-N and FIX-R mean the normal cell from GDAS (FRC) and our designed reduction cell, {respectively}.
}
\label{table:ablative}
\end{table}

\textbf{Results on WT2}.
To train the model on WT2, we use the same experiment settings as PTB, but we use a hidden size of 700 and a weight decay of 5e-7. We train the model in 3000 epochs in total.
\Tabref{table:WT2} compares different RNN models on WT2.
Our approach achieves competitive results among all automatically searching approaches.
GDAS is worse than ``LSTM + SC''~\cite{merity2018regularizing}.
Since our model is searched on a small dataset PTB, and the transferable ability of the discovered model might be a little bit weak. If we directly search the RNN model on WT2, we could obtain a better model and improve the transferable ability.

\begin{figure}[t!]
\begin{center}
\includegraphics[width=\linewidth]{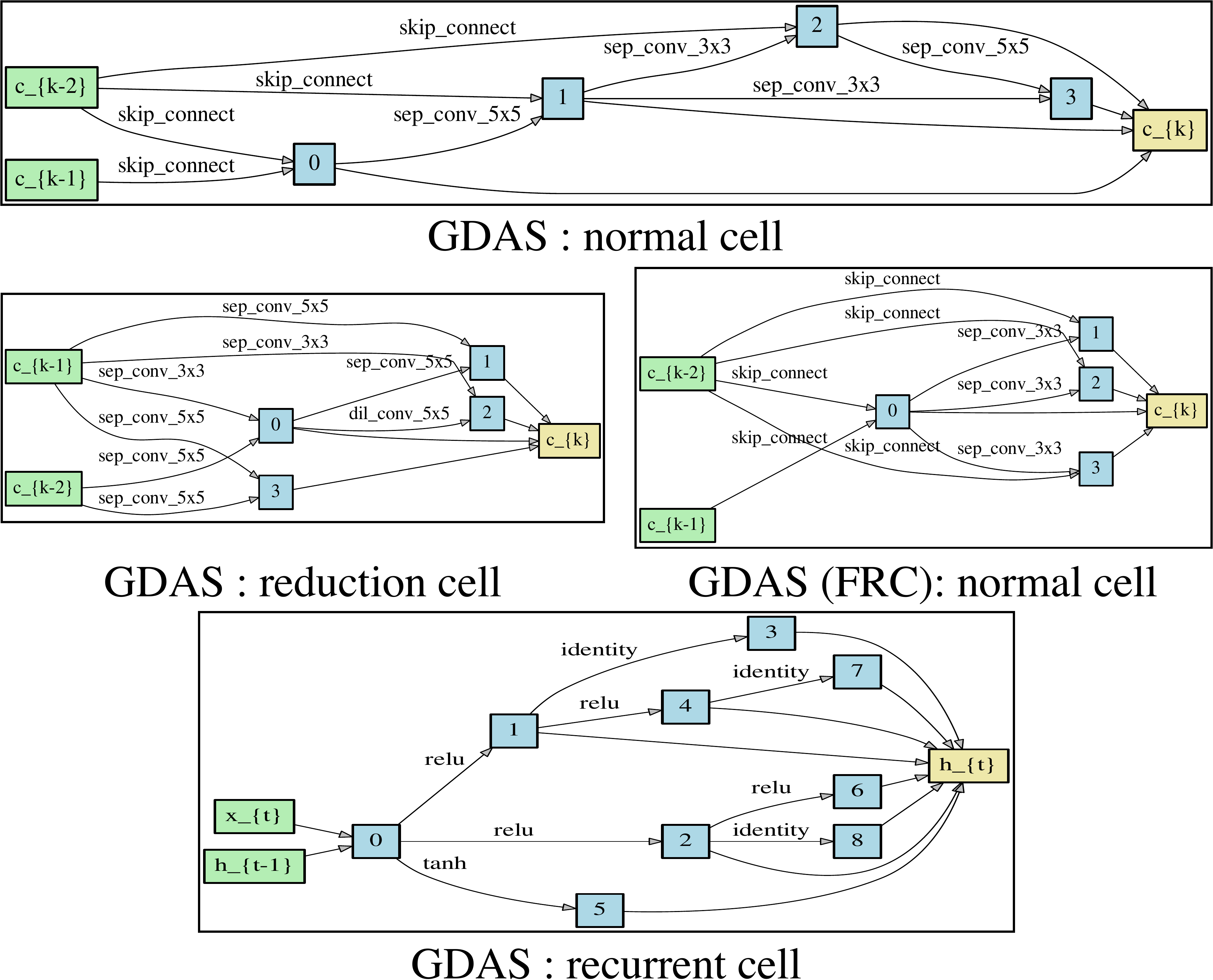}
\end{center}
\caption[Captioning]{
The top block and the middle-left block show the normal cell and the reduction cell discovered by GDAS, {respectively}. The middle-right block shows the discovered normal cell when we fix the reduction cell. The bottom block shows the discovered recurrent cell.
}
\vspace{-2mm}
\label{fig:cell-vis}
\end{figure}

\subsection{Discussion}\label{sec:exp-dis}

We visualize the discovered cells in \Figref{fig:cell-vis}.
These automatically discovered cells are complex and hard to be designed manually.
Moreover, networks with these discovered cells can achieve more superior performance than hand-crafted networks.
This demonstrates that automated neural architecture search is the future of architecture design.

Revisiting \Secref{sec:method-reduction}, we propose a new reduction cell as a replacement for automated reduction cell.
With this reduction cell, we can more effectively search neural cells.
For further analysis, we use the normal cell found by GDAS and the proposed reduction cell to construct a new CNN, denoted as ``GDAS-N + FIX-R'' in \Tabref{table:ablative}.
The accuracy of this network on CIFAR-10 is similar to ``GDAS-N + GDAS-R'' and ``FRC-N + FIX-R'' in \Tabref{table:ablative}.
This result implies that the reduction cell might have a negligible effect on the performance of networks and the hand-crafted reduction cell could be on par with the automatically discovered one.

Most recent NAS approaches search neural networks on the small-scale datasets, such as CIFAR, and then transfer the discovered networks to the large-scale datasets, such as ImageNet. The obstacle of directly searching on ImageNet is the huge computational cost. GDAS is an efficient NAS algorithm and gives us an opportunity to search on ImageNet. We will explore this research direction in our future work.

\section{Conclusion}

In this paper, we propose a Gradient-based neural architecture search approach using Differentiable Architecture Sampler (GDAS).
Our approach is efficient and reduces the search cost of the standard NAS approach~\cite{Zoph_2018_CVPR} by about 10$^{4}$ times.
Moreover, both CNN and RNN models discovered by our GDAS can achieve competitive performance compared to state-of-the-art models.

{\small
\bibliographystyle{ieee_fullname}
\bibliography{egbib}
}

\end{document}